\documentclass[nohyperref]{article}

\usepackage{microtype}
\usepackage{graphicx}
\usepackage{subfigure}
\usepackage{bbm}
\usepackage{booktabs} 

\usepackage{hyperref}
\usepackage{tabularx}

\newcommand{\smallsec}[1]{\vspace{0.2em}\noindent\textbf{#1}}

\usepackage[accepted]{icml2022}

\usepackage{amsmath}
\usepackage{amssymb}
\usepackage{mathtools}
\usepackage{amsthm}

\usepackage[capitalize,noabbrev]{cleveref}

\usepackage{wrapfig}
\usepackage{listings}
\usepackage{algorithm}
\usepackage{pifont}

\usepackage{booktabs}    
\usepackage{arydshln}
\usepackage{mathtools}

\newcolumntype{L}[1]{>{\raggedright\let\newline\\\arraybackslash\hspace{0pt}}m{#1}}
\newcolumntype{C}[1]{>{\centering\let\newline\\\arraybackslash\hspace{0pt}}m{#1}}
\newcolumntype{R}[1]{>{\raggedleft\let\newline\\\arraybackslash\hspace{0pt}}m{#1}}

\theoremstyle{plain}

\theoremstyle{definition}

\theoremstyle{remark}

\newcommand{\xmark}{\ding{55}}%
\newcommand\numberthis{\addtocounter{equation}{1}\tag{\theequation}}

\usepackage[disable,textsize=tiny]{todonotes}

\icmltitlerunning{Object Permanence Emerges in a Random Walk along Memory}

\begin{document}

\twocolumn[
\icmltitle{Object Permanence Emerges in a Random Walk along Memory}



\icmlsetsymbol{equal}{*}

\begin{icmlauthorlist}
\icmlauthor{Pavel Tokmakov}{tri}
\icmlauthor{Allan Jabri}{berk}
\icmlauthor{Jie Li}{tri}
\icmlauthor{Adrien Gaidon}{tri}
\end{icmlauthorlist}

\icmlaffiliation{tri}{Toyota Research Institute}
\icmlaffiliation{berk}{UC Berkeley}

\icmlcorrespondingauthor{Pavel Tokmakov}{pavel.tokmakov@tri.global}

\icmlkeywords{Machine Learning, ICML}

\vskip 0.3in
]


\printAffiliationsAndNotice{}  

\begin{abstract}
This paper proposes a self-supervised objective for learning representations
that localize objects under occlusion - a property known as object permanence.
A central question is the choice of learning signal in cases of total occlusion.
Rather than directly supervising the locations of invisible objects, 
we propose a self-supervised objective that requires neither human annotation, nor assumptions about object dynamics. We show that object permanence can emerge by optimizing for temporal coherence of memory: we fit a Markov walk along a space-time graph of memories, where the states in each time step are non-Markovian features from a sequence encoder.
This leads to a memory representation that stores occluded objects and predicts their motion, to better localize them.
The resulting model outperforms existing approaches on several datasets of increasing complexity and realism, despite requiring minimal supervision, and hence being broadly applicable.
\end{abstract}

\section{Introduction}
\label{sec:intro}
\begin{figure}[t]
\begin{center}
  \includegraphics[width=0.8\columnwidth]{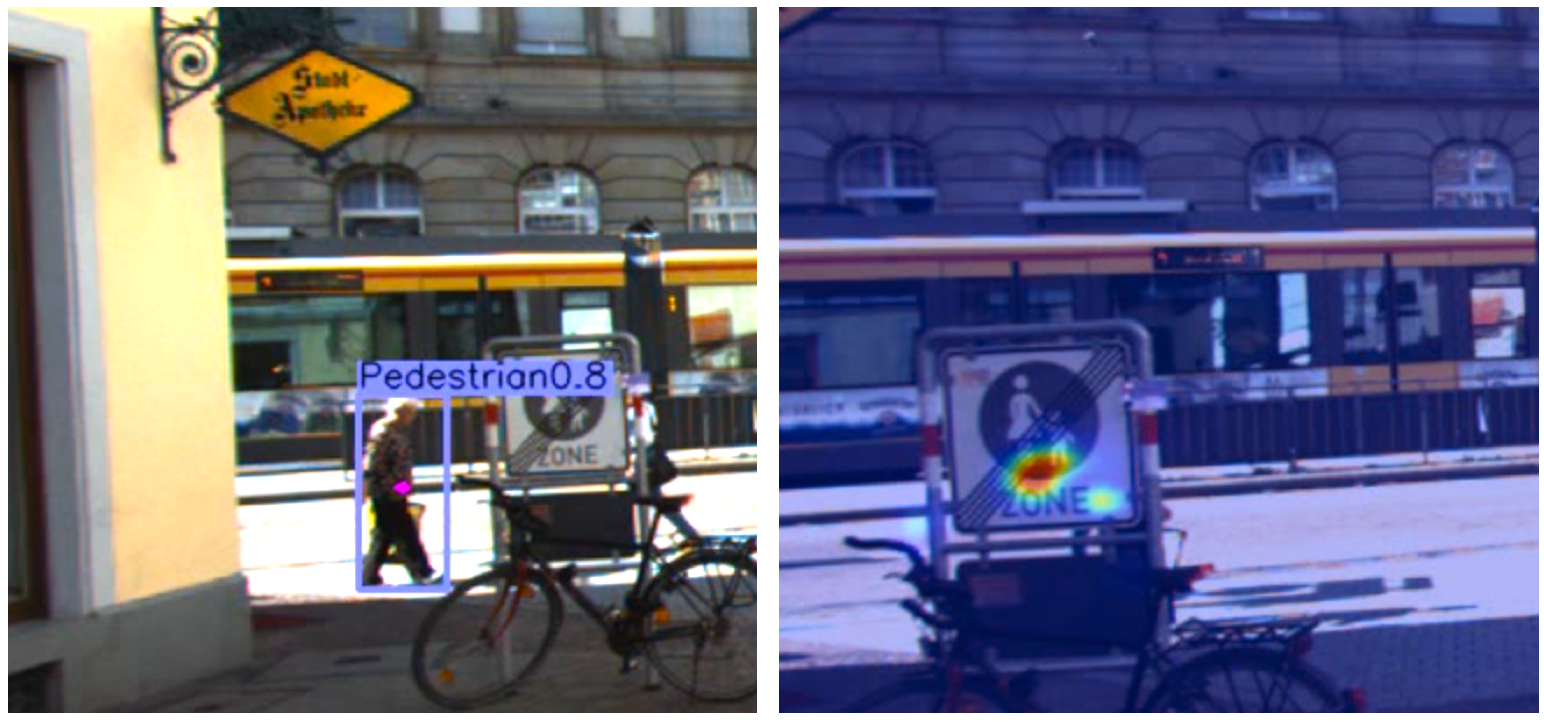}
\end{center}
\vspace*{-2mm}
  \caption{
  An example from the KITTI dataset with outputs of our method (object detection on the left and belief state on the right). We demonstrate that by fitting a Markov walk along an evolving spatial memory, a spatial belief state that codes object permanence emerges without explicit supervision.
  }
\vspace{-5mm}
\label{fig:teaser}
\end{figure}

Object permanence -- the notion that objects, such as the person in Figure~\ref{fig:teaser}, continue to exist even when occluded -- is a crucial component of perception. It is fundamental in development~\cite{baillargeon1985object,spelke1990principles}, and critical for perception and control in partially observable environments like the physical world~\cite{kaebling98, grabner2010tracking, schmidt2014dart, garg2021semantics, tokmakov2021learning}.
Yet, modern machine learning models for object recognition are mostly limited by instantaneous observations and struggle with occlusions~\cite{he2017mask,zhou2019objects}. 
Recent video-based methods~\cite{xiao2018video,shamsian2020learning,tokmakov2021learning} have the capacity to localize fully invisible instances by modeling sequential structure. For example, Tokmakov et al.~\yrcite{tokmakov2021learning} use a spatial recurrent network~\cite{ballas2015delving} to accumulate a representation of a scene and localize instances -- both visible and invisible -- using this representation.
Nevertheless, a key question that remains is the choice of learning signal in cases of total occlusion. 

While Shamsian et al.~\yrcite{shamsian2020learning} propose to supervise the model with the ground truth locations of invisible instances, such labels are hard to obtain in the real world and often sub-optimal~\cite{tokmakov2021learning}. For example, consider the sequence in Figure~\ref{fig:teaser}, in which a person walks behind a street sign. Without additional observations, it is impossible to predict their \textit{exact} location, even for a human, let alone for a machine learning model, only that they are somewhere behind that sign.

\begin{figure}[t]
\begin{center}
  \includegraphics[width=0.9\columnwidth]{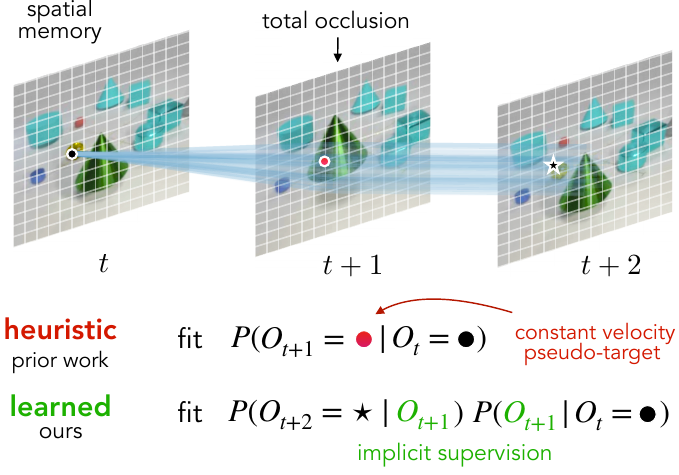}
\end{center}
\vspace*{-4mm}
  \caption{
  How to model object location under total occlusion? While prior work considers heuristic dynamics priors such as constant velocity, we propose to learn priors from data. Instead of relying on pseudo-targets, we enforce a Markov assumption on the spatial memory representation, implicitly supervising multiple hypotheses about the occluded object location.
}
\vspace{-6mm}
\label{fig:teaser2}
\end{figure}

Rather than directly supervising the locations of invisible objects, in this work we propose a self-supervised objective that encourages object permanence to naturally emerge from data (see Figure~\ref{fig:teaser2}). To this end, we leverage the recent Contrastive Random Walk objective of Jabri et al.~\yrcite{jabri2020space}, which models space-time correspondence as a Markov walk on a spatio-temporal graph of patches (i.e. from a video). Instead of supervising the walker at each step, which requires temporally dense annotation, they supervise every $k$ steps,  providing implicit supervision for the trajectory. 
Our key insight is that object permanence emerges by fitting such a Markov walker \textit{along an evolving spatial memory}, provided that the states in each time step are features produced by a sequence encoder, so as to overcome partial observability. 

We propose a self-supervised loss function that can be applied to any video object detection model that maintains a spatial memory (i.e. a sequence of 2D latents). Concretely, we consider the spatio-temporal graph of memory states, where nodes correspond to potential object locations 
(Figure~\ref{fig:arch}). For visible instances, transition probability is supervised directly (i.e. we assume labels for \textit{visible} objects during training). During occlusions, we employ the objective of Jabri et al.~\yrcite{jabri2020space}, supervising the walk with ground truth object locations before and after the occlusion (see Figure~\ref{fig:method}). 
By optimizing for correspondence on the resulting graph of memories, we learn a representation that stores object-centric information in a spatially-grounded manner even for unlabeled, invisible objects.

We demonstrate that object permanence naturally emerges in this process, evaluating our approach on several dataset of increasing complexity and realism (see Figures~\ref{fig:lacater}, \ref{fig:real}). We begin with the synthetic LA-CATER benchmark~\cite{shamsian2020learning} for invisible object localization on which our self-supervised objective is able to discover object permanence patterns from the data, outperforming a fully-supervised baseline. We then evaluate our method on a synthetic multi-object tracking benchmark introduced  in~\cite{tokmakov2021learning}, and demonstrate that it is effective at handling occlusions despite requiring less supervision. Finally, we show that our method generalizes to real world videos in the KITTI multi-object tracking benchmark~\cite{geiger2012we}, and provide a detailed ablation analysis. Source code, models, and data are publicly available at \url{https://tri-ml.github.io/RAM}.

\section{Related Work}
Our work addresses the problem of localizing and associating occluded objects (commonly referred to as \textit{object permanence}) in a \textit{spatio-temporal video representation} via \textit{self-supervised learning of correspondence}. Below, we review the most relevant approaches in each of these fields.

\vspace{-4mm}
\paragraph{Object permanence} has been mostly studied in the context of multi-object tracking~\cite{luo2020multiple}, where localizing invisible objects is crucial for re-associating them after the occlusion. The majority of the modern trackers operate in the tracking-by-detection paradigm~\cite{bewley2016simple,wojke2017simple}, where objects are first localized with a frame-level detector, and the resulting detections are then associated based on bounding box overlap~\cite{bewley2016simple}, or feature similarity~\cite{wojke2017simple,tang2017multiple,xu2019spatial}.

A key limitation of these approaches is that a frame-level detector can only localize visible instances, thus most of them had to resort to heuristics to model object permanence. Some of the early methods include~\cite{huang2005tracking,papadourakis2010multiple}, which attempted to localize occluded objects by modeling inter-occlusion relationships, and the approach of Grabner et al.~\yrcite{grabner2010tracking}, which captured the correlation between the motion of visible and invisible instances. However, in practice most methods relied on a more robust constant velocity heuristic~\cite{yu2007multiple,breitenstein2009robust,mitzel2010multi}, which propagates the last observed object location with a linear motion model in the frame coordinates.

Recently, Shamsian et al.~\yrcite{shamsian2020learning} proposed a learning-based method which takes bounding boxes for visible objects in a video as input, and passes them through a sequence of LSTMs~\cite{hochreiter1997long} that are trained to localize the occluded object. This approach successfully learns to capture complex relationship between visible and occluded instances. However, it requires ground truth labels for invisible objects for training and relies on the assumption that the location of an occluded object is fully determined by the bounding box of the occluder. In contrast, our method does not require any labels for invisible objects, does not make assumptions about their dynamics, and still outperforms the approach of Shamsian et al.~\yrcite{shamsian2020learning}.
The method of Tokmakov et al.~\yrcite{tokmakov2021learning} also learns to localize occluded objects, but uses a spatio-temporal representation. 
\vspace{-4mm}
\paragraph{Spatio-temporal video representation} learning allows to jointly model visible and occluded instances by capturing temporal context. Several methods~\cite{feichtenhofer2017detect,bergmann2019tracking,zhou2020tracking} operate on frame pairs to improve robustness, but are neither capable of, nor trained for handling full occlusions. Some video object detection~\cite{kang2017object,xiao2018video} and segmentation~\cite{bertasius2020classifying,wang2021end} methods take longer sequences as input, but are also not trained to localize invisible objects.

Very recently, Tokmakov et al.~\yrcite{tokmakov2021learning} extended the model of Zhou et al.~\yrcite{zhou2020tracking} to videos of arbitrary length. In particular, they used a Convolutional Gated Recurrent Unit (ConvGRU)~\cite{ballas2015delving} to aggregate a spatial memory representation that is capable of encoding both visible and invisible objects. Following~\cite{zhou2020tracking}, they then represented objects with their centers in the frame coordinates and trained the model to localize and associate them using the memory state regardless of visibility. 

The key question is the choice of the learning signal for fully occluded instances. The authors of~\cite{tokmakov2021learning} trained their model on synthetic videos, where ground truth labels for invisible objects are available, but found that this form of supervision is sub-optimal as the exact location of an invisible object is often impossible to predict. Instead, they used deterministic pseudo-groundtruth obtained by propagating object location in the 3D world with a constant velocity and projecting it to the camera frame. While this approach simplifies convergence, the resulting model often fails at test time when object behaviour deviates significantly from the constant velocity. In this work, we adopt the architecture proposed in~\cite{tokmakov2021learning}, but demonstrate that occluded object localization can be learned without direct supervision by optimizing for a Markov walk along an evolving spatial memory state. Our self-supervised objective outperforms the method of Tokmakov et al.~\yrcite{tokmakov2021learning} on both synthetics and real-world videos.

\vspace{-4mm}
\paragraph{Self-supervised learning of motion correspondence} has emerged recently as a way to capitalize on temporal consistency in videos for representation learning~\cite{wang2019learning,wang2019unsupervised,jabri2020space}. The key idea is to utilize cycle-consistency in time~\cite{zhou2016learning,dwibedi2019temporal} to learn to establish correspondences between patches in consecutive frames. In particular, given a randomly selected patch in the first frame, these methods first track forward in time, then backward, with the aim of ending up where they started. Earlier approaches~\cite{wang2019learning,wang2019unsupervised} relied on hard attention, limiting them to sampling and learning from one path at a time. Recently~\cite{jabri2020space} have proposed the Contrastive Random Walk objective which computes soft-attention at every time step, considering many paths to obtain a dense learning signal. 

While these objectives are beneficial for general representation learning, they have found wider success in video object segmentation~\cite{pont20172017}, where a region in the first frame needs to be propagated through the video. However, since these methods operate on individual image encodings, they can only establish correspondences based on appearance. Thus, similarly to tracking-by-detection approaches described above, they can not handle occlusions. In this work, we adapt the objective of Jabri et al.~\yrcite{jabri2020space} to learn to localize occluded objects in a self-supervised way. We apply it to sequence-level representations \cite{tokmakov2021learning} in order to enforce temporal coherence of spatial memory, even during occlusion, and show that object permanence naturally emerges in this process.

\section{Approach}
\label{sec:meth}
\subsection{Preliminaries}
We study the problem of localizing entities as they transform in space and time, both when they are visible and after they become fully occluded. Given a  sequence of frames $\{I^{1},I^2, ..., I^n\}$, we follow the formalism recently proposed in~\cite{zhou2019objects,zhou2020tracking} and model objects $o_i^t$ with their centers in image coordinates  $\mathbf{p}_i^t \in \mathbb{R}^2$.

We consider models that maintain a spatial memory $M^t \in \mathbb{R}^{D \times H' \times W'}$. 
Typically, images $I^t \in \mathbb{R}^{3 \times H \times W}$ are mapped by an encoder $f$ to feature maps $F^t = f(I^t)$. These instantaneous observations are then aggregated using a sequence model. While this can be achieved in a variety of manners, including encoders with global self-attention~\cite{vaswani2017, bertasius2021}, we consider the ConvGRU~\cite{ballas2015delving}, a spatial recurrent network that is efficient, performant, and online: $M^t=\mathtt{ConvGRU}(F^t, M^{t-1})$,
where $M^t, M^{t-1}$ represent the current and the previous spatial memory states respectively. The state $M^t$ is thus informed by prior context of extant objects when integrating updates $F^t$ from the current frame, and can encode the locations of both visible and invisible objects.

However, the choice of the learning signal for the cases of total occlusion remains a central question. In~\cite{tokmakov2021learning} the authors propose to treat visible and invisible instances uniformly, directly supervising their center locations $\{\mathbf{p}_1^t, \mathbf{p}_2^t, ..., \mathbf{p}_N^t\}$ on the learned projection of the spatial memory $P^t = f_{\mathbf{p}}(M^t) \in [0, 1]^{H' \times W'}$.
We adopt this approach for visible object, but propose a novel self-supervised framework for learning to localize the invisible ones in the next section.
It is based on a Contrastive Random Walk~\cite{jabri2020space} along the memory state $M^t$ and learns to estimate the locations of occluded object centers without explicit supervision.
\begin{figure}[t]
\begin{center}
  \includegraphics[width=0.85\columnwidth]{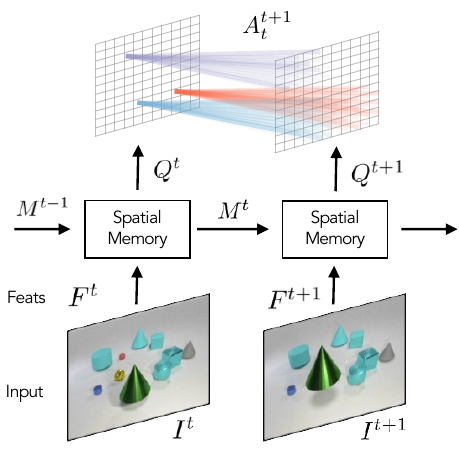}
\end{center}
\vspace*{-4mm}
  \caption{We consider the spatio-temporal graph of an evolving spatial memory; here, we show one transition in time. To overcome partial observability, states $Q^t$ are computed with a sequence encoder, allowing for transition probability $A_t^{t+1}$ to model object permanence. Only a subset of the edges is shown for readability. }
\vspace*{-4mm}
\label{fig:arch}
\end{figure}

\begin{figure*}
    \centering
    \hspace{-2mm}
    \includegraphics[width=1\linewidth]{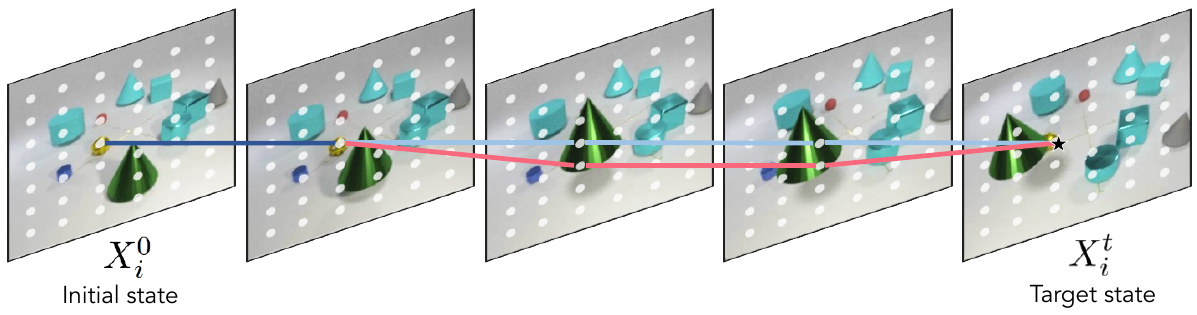}
    \vspace*{-4mm}
    \caption{Illustration of our objective on a sequence from LA-CATER. We initiate a random walk on the graph originating from a visible object center. While the object remains visible the walker state is supervised directly. During occlusions the walker is free to take any path as long as it terminates at the object center at the time of disoclusion (shown with a star in the last frame). Multiple hypothesis about the object trajectory (shown in blue and red) are thus implicitly supervised.}
    \label{fig:method}
    \vspace{-4mm}
\end{figure*}

\subsection{Walking along Memory}
The main idea is to learn object permanence by fitting a \textbf{r}andom walk \textbf{a}long \textbf{m}emory (RAM). We construct a spatio-temporal graph over the memory state, shown in Figure~\ref{fig:arch}, with pixels on the feature map $Q^t = f_{\mathbf{q}}(M^{t})$ as nodes $\{\mathbf{q}_1^t, \mathbf{q}_2^t, ..., \mathbf{q}_m^t\}$. Only nodes in consecutive frames $Q^t, Q^{t+1}$ are sharing an edge. The strength of an edge is determined by the similarity of the node embeddings $d(\mathbf{q}_i^t, \mathbf{q}_j^{t+1})=\langle{}\mathbf{q}_i^t, \mathbf{q}_j^{t+1}\rangle{}$, which is converted into non-negative affinities by applying a softmax over edges originating from each node:
\vspace{-2mm}
\begin{align*}
    A_t^{t+1}(i,j) &= \mathtt{softmax} (Q^t,Q^{t+1})_{ij} \\
    & =  \frac{\exp({d(\mathbf{q}_i^t, \mathbf{q}_j^{t+1})/\tau})}{\sum_{l=1}^N \exp({d(\mathbf{q}_i^t, \mathbf{q}_l^{t+1})/\tau})} \numberthis 
\label{eq:edge}
\end{align*}
where $\tau$ is the temperature parameter. In contrast to~\cite{jabri2020space}, in this work we build the graph over the evolving memory $M^t$, not over independently encoded features $F^t$. As a result, the nodes can represent invisible objects and the transition probability is not solely determined by similarity of instantaneous appearance.

\textbf{Fitting the walk.} For each training sequence, the model receives a set of objects annotations $\{O_1, O_2, ..., O_N\}$ as input, where an object is represented with its \textit{visible} bounding box centers $O_i = \{\mathbf{p}_i^0, \mathbf{p}_i^1, \emptyset, \emptyset, ...,  \mathbf{p}_i^t, ..., \mathbf{p}_i^T\}$, and empty annotations $\emptyset$ correspond to frames in which the object is occluded. For each object $O_i$ we consider the random walk originating from the first visible object center $\mathbf{p}_i^0$ (shown in Figure~\ref{fig:method}; we assume the object is visible in the first frame of the sequence). In particular, we initialize the walker state matrix $X_i^0$ with $1$ at $\mathbf{p}_i^0$ and 0 everywhere else, and compute the distribution of the object location at time $t$ as
\begin{equation}
     X_i^{t} = X_i^0 \prod_{j=0}^{t-1} A_{j}^{j + 1}.
\label{eq:step}
\end{equation}
That is, $X_i^{t}$ is a conditional probability matrix with $X_i^{t}(\mathbf{p})=\mathbb{P}(O_i^t=\mathbf{p} | O_i^0)$ representing the probability that object $i$ is at position $\mathbf{p}$ at time $t$, given its initial position $\mathbf{p}_i^0$.
Ground truth boxes of visible objects supervise the walker via loss
\begin{equation}
    L_{NLL}(X_i^{t}, \mathbf{p}_i^{t}) = - \log X_i^{t}(\mathbf{p}_i^t),
\label{eq:ce}
\end{equation}
where $L_{NLL}$ is the negative log likelihood of the correct position. The total loss for the object $O_i$ is defined as 
\begin{equation}
    L_{RAM}(O_i) = \sum_{t=1}^{T} \mathbbm{1}(\mathbf{p}_i^{t}) L_{NLL}(X_i^{t}, \mathbf{p}_i^{t}),
\label{eq:crw}
\end{equation}
where $\mathbbm{1}(\mathbf{p}_i^{t})$ is the indicator function which is equal to $1$ for non-empty object center labels $\mathbf{p}_i^{t}$ and is $0$ otherwise. The final objective is averaged over all the instances in the scene: $L_{RAM} = \frac{1}{N} \sum_{i=1}^{N} L_{RAM}(O_i)$.

The loss above directly supervises the object centers in frames in which the object is visible (as in~\cite{tokmakov2021learning}). In cases of occlusion, there are many potential paths through the graph that link the object's locations before and after occlusion. Minimizing the RAM objective in Equation~\ref{eq:crw} shifts the probabilities towards the paths which are most likely to result in correctly localizing the object when it re-appears  (shown in blue and red in Figure~\ref{fig:method}).
The locations of invisible objects are thus implicitly supervised without the need for any labels and with minimal assumptions about dynamics. The resulting encoder learns to store the spatially-grounded object-centric information in memory $M^t$ to guide the walker along typical trajectories. Next, we discuss a more efficient variant of this approach that exploits smoothness and locality.

\subsection{Local Attention on Memory} 
\label{seq:lrw}
Notice that computing the edge weights matrix $A_t^{t+1} \in  \mathbb{R}^{H'W' \times H'W'}$ is the most computationally and memory intensive operation in the RAM objective. It requires $D \times H'W' \times H'W'$ multiplications and the size of the matrix can get prohibitively large for large resolution frames. To mitigate this constraint, we exploit smoothness in videos and assume that between any pair of consecutive frames $I_t, I_{t+1}$ the pixels can only shift by a limited distance $r$. 

Given a node $\mathbf{q}_i^t$ with coordinates $\mathbf{p}_i^{t}$, its corresponding row in the (now sparse) matrix $A_t^{*t+1}(i, :)$ is:
\begin{equation}
A_t^{*t+1}(i, j) =
\begin{cases}
  A_t^{t+1}(i, j), & \text{if }
       \begin{aligned}[t]
       || \mathbf{p}_i^{t} - \mathbf{p}_j^{t+1} ||_1 < r
       \end{aligned}
\\
  \emptyset, & \text{otherwise},
\end{cases}
\label{eq:local}
\end{equation}
where $A_t^{t+1}(i, j)$ is defined in Equation~\ref{eq:edge}, with softmax applied over non-zero edges only. That is, we only compute local attention between $\mathbf{q}_i^t$ and the nodes in its neighbourhood in $Q^{t+1}$, which can be efficiently implemented with local correlation operations~\cite{dosovitskiy2015flownet}. As we will see, with a sufficiently large $r$ this modification does not limit the performance of the approach, while reducing its computational complexity. 
In the next section, we conclude by discussing the final details of our objective.

\subsection{Optimization and Objective} 
\label{seq:ws}
So far, for any given walker state $X_i$ we have only constrained it at frames in which the corresponding object is visible. We now demonstrate how additional information about the problem can be incorporated into the state to simplify convergence and improve tracking performance.

\vspace{-2mm}
\paragraph{Label smoothing.} Node features $\mathbf{q}_i^t$ are extracted from a spatial memory state computed with a convolutional backbone, and nodes that are spatially close to each other have highly correlated representations. Yet, minimizing the contrastive objective in Equation~\ref{eq:crw} encourages the transition matrix $A_{t-1}^t(i, :)$ to have low entropy, such that nodes in the direct vicinity of the ground truth node $\mathbf{q}_i^t$ are very hard negatives that may hinder optimization. 

To mitigate this issue, we relax the objective by scaling the loss applied on state matrix $X_i^t$ to ignore transitions to nodes around the ground-truth object center. Concretely, we multiply $X_i^t$ by a mask $H_i^t \in [0, 1]^{H' \times W'}$ before passing it to the loss in Equation~\ref{eq:crw}. The mask is computed according to~\cite{zhou2019objects}  and decreases the walker probability for the nodes within the immediate vicinity of $\mathbf{q}_i^t$. This amounts to reducing their effect as negatives, akin to label smoothing around the center.

\vspace{-2mm}
\paragraph{Avoiding overlap.} At inference time, we follow~\cite{zhou2020tracking} and associate objects based on distances between their centers in consecutive frames. For an occluded object $O_i$
we estimate its object center at time $t+ k$ as
\begin{equation}
\hat{\mathbf{p}}_i^{t+k} = \underset{\mathbf{p}}{\arg\max}~X_i^{t+k}(\mathbf{p})
\end{equation}
and match it to the centers of the detected boxes in that frame (see Section~\ref{ap:tracking} for details). If a match is found, we assume the object has re-appeared and terminate the walk. However, in practice the estimated center $\hat{\mathbf{p}}_i^{t+k}$ might overlap with the center of the occluder, leading to an incorrect association. To avoid this, during training we penalize the walker state of an occluded object if it overlaps with visible centers:
\begin{equation}
    L_{over} (X_i^t, \mathbf{p}_:^{t}) = \sum_{j \neq i} X_i^t(\mathbf{p}_j^t),
\end{equation}
where $X_i^t(\mathbf{p}_j^t)$ is the value of the walker state at the location corresponding to the ground truth center of the visible object $j$. Note that $\mathbf{p}_j^t$ might in fact be the correct location for the center of the occluded object $i$, but avoiding this point in the frame space allows us to prevent an incorrect association. To fully address the center overlap issue frame representations would have to be mapped to the BEV space, which is a major challenge. Our approach allows for a simple and effective (even if not principled) way to avoid overlaps, while remaining in the image plane.

\vspace{-2mm}
\paragraph{Objective.} The overall training objective is then defined as follows:
\begin{equation}
    L = L_{PT} + \lambda_{1} L_{RAM} + \lambda_{2}L_{over},
\label{eq:loss}
\end{equation}
where $L_{PT}$ represents the losses adopted from PermaTrack~\cite{tokmakov2021learning} for localization and association of the visible objects, and $\lambda_{1}, \lambda_{2}$ are hyper-parameters that balance the contribution of the corresponding losses in the overall objective.

\subsection{Inference Algorithm}
\label{ap:tracking}
To conclude, we describe the algorithm used in our work to localize occluded object at test time. Overall, we follow the greedy association strategy proposed in~\cite{zhou2020tracking}, which maintains a history of the locations of previously seen objects and matches them with the new detections in frame $t$ based on center distances. Unmatched tracks then corresponds to objects for which no detection could be associated, usually due to an occlusion. We summarize our approach for handling such trajectories in Algorithm~\ref{alg:infer}. 

We begin by initializing a walker state $X_i^{t-1}$ with the last observed object center $\hat{\mathbf{p}}_i^{t-1}$ for each occluded object individually. We then update the walker states with the transition probability matrix $A_{t-1}^t$ and obtain the maximum likelihood hypothesis of each occluded object location by taking argmax of the resulting distribution $X_i^{t}$. Note that the transition matrix is shared between the walkers, and only individual states need to be maintained, improving scalability of the approach. Center location hypotheses are used to match the trajectories to the detections in the current frame with a greedy strategy similar to~\cite{zhou2020tracking}. If no match is found, the walker continues forward in the same way, and is terminated if its confidence falls below a threshold {\tt conf\_th} or the trajectory goes out of frame.

\begin{algorithm}[H]
 \caption{\small A step of the inference algorithm in Python style. }
        \label{alg:infer}
        \definecolor{codeblue}{rgb}{0.25,0.5,0.5}
        \lstset{
          backgroundcolor=\color{white},
          basicstyle=\fontsize{7.2pt}{7.2pt}\ttfamily\selectfont,
          columns=fullflexible,
          breaklines=true,
          captionpos=b,
          commentstyle=\fontsize{7.2pt}{7.2pt}\color{codeblue},
          keywordstyle=\fontsize{7.2pt}{7.2pt},
          escapechar=\&
        }
\begin{lstlisting}[language=python]
for t in unmatched_tracks:  # t: track hypothesis
  if "walk" not in t:
    #occlusion start, initialize the walker state 
    t["walk"] = init_walk(t["center"])

  #update the walker with transition probabilities A
  t["walk"] = t["walk"] * A
    
  #Maximum likelihood confidence and location
  conf, ind = t["walk"].max()
  
  #drop unreliable hypotheses
  if conf < conf_th or at_boundary(ind):
    continue
    
  #greedy matching with detections
  is_matched, det = match(ind, detections)
  if is_matched:
    #use visible object location if a match is found
    t["center"] = det["center"]
  else:
    #otherwise store walker hypothesis
    t["center"] = ind

  #add track to the active pool
  tracks += t &\hspace{5mm}& 
\end{lstlisting}
\end{algorithm}

\section{Experimental Evaluation}

\subsection{Datasets and Evaluation}
We use three datasets of increasing complexity and realism: a synthetic LA-CATER benchmark~\cite{shamsian2020learning}, a photo-realistic synthetics PD dataset~\cite{tokmakov2021learning}, and a real-world, multi-object tracking KITTI dataset~\cite{geiger2012we}. Below, we describe each of these benchmarks in more detail together with their metrics. 

\begin{figure*}
    \centering
    \vspace{-3mm}
    \includegraphics[width=0.95\linewidth, height=3.2in]{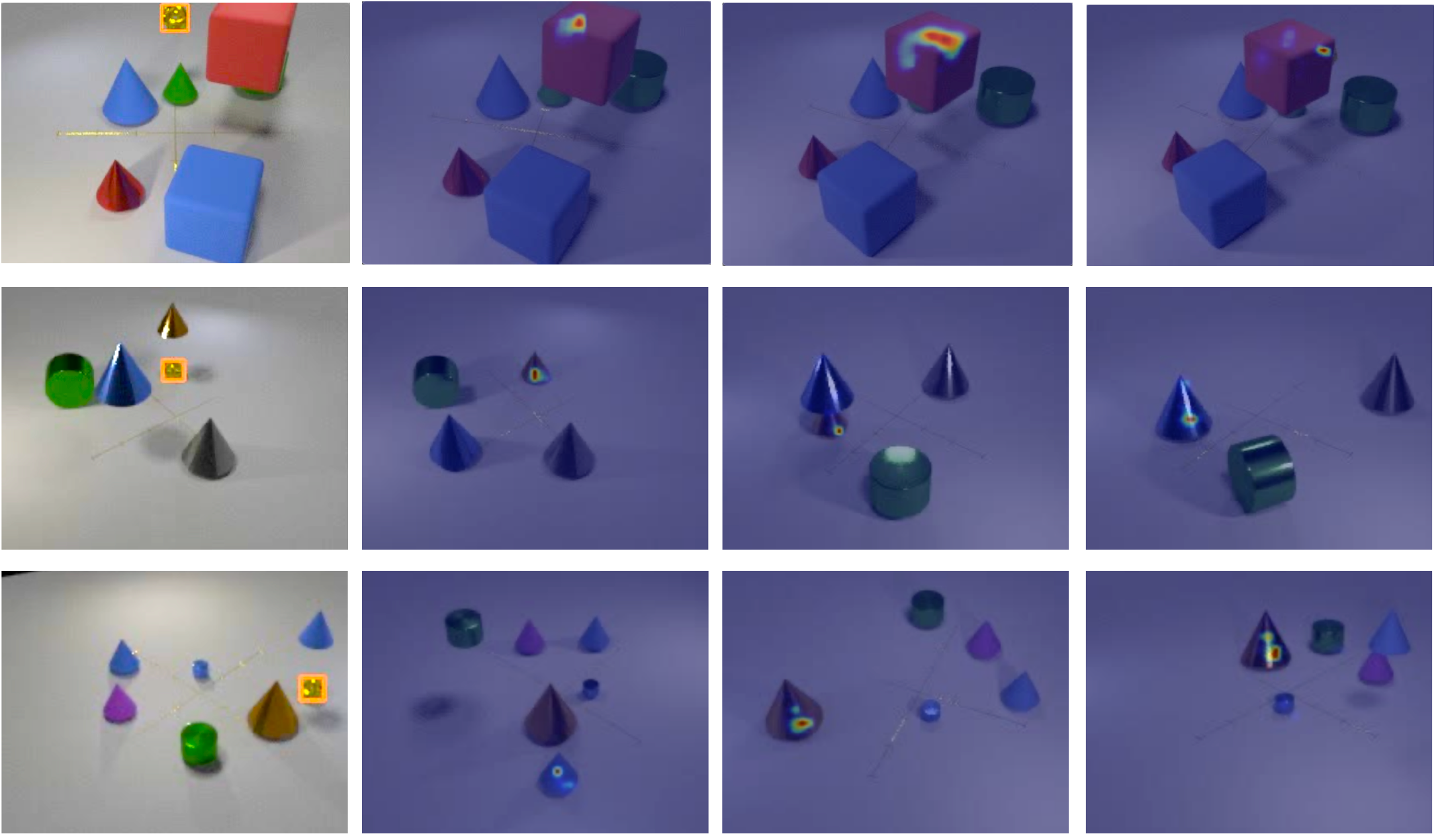}
    \vspace{-4mm}
    \caption{Qualitative results on sequences from the test set of LA-CATER-Moving (see \href{https://youtu.be/W9lsSG55Xzw}{link} for full results). The model's belief about the location of the invisible object is visualized with a heatmap overlaid on the frames. Our approach successfully handles examples of occlusion (top), containment (middle), and carrying (bottom) without using any invisible object labels for training.}
    \vspace{-4mm}
    \label{fig:lacater}
\end{figure*}

\vspace{-2mm}
\paragraph{LA-CATER} is based on the CATER dataset~\cite{girdhar2019cater} which is procedurally generated using the Blender 3D engine~\cite{bacone2012blender}. The generated scenes consist of a random number of geometric shapes (cube, sphere, cylinder, or cone) placed on a plain background. An additional golden sphere is added to every scene. All the objects are set to move randomly, occluding each other and the sphere (see Figure~\ref{fig:lacater}). A special form of occlusion - containment, is introduced by cones covering the sphere.  

In~\cite{shamsian2020learning}, the authors converted CATER to an object permanence benchmark by generating ground truth boxes for all the objects and labeling each frame with the state of the golden sphere (visible, occluded, contained, or carried). Containment corresponds to it being fully covered by another object, and carried frames are those in which the container is moving with the sphere underneath. The task is then to localize the golden sphere in every frame. Each video is 10-seconds long. There are 9300 training, 3327 validation and 1371 test videos respectively. Performance is measured with intersection over union (IoU) between ground-truth and predicted boxes~\cite{everingham2010pascal}. 

In LA-CATER, the camera position is fixed for all the videos, making the problem less challenging. In this work, we generated a variant of the dataset with randomized camera motion, which we refer to as LA-CATER-Moving. This benchmark has the same statistics and annotations as the original LA-CATER, allowing us to re-train the baselines from~\cite{shamsian2020learning} and compare to their method in a more realistic environment.

\vspace{-3mm}
\paragraph{PD} dataset is collected by Tokmakov et al.~\yrcite{tokmakov2021learning} using a state-of-the-art ParallelDomain synthetic data generation service~\cite{parallel_domain}. The dataset contains 210 photo-realistic, 10-seconds long videos with driving scenarios in city environments captured at 20 FPS. Each scene contains dozens of objects, including pedestrians, cars, bicycles, etc., and features lots of occlusion and disocclusion scenarios (only people and cars are used for evaluation). The videos are captured by three independent cameras, effectively increasing the dataset size to 630. Following~\cite{tokmakov2021learning}, we use $583$ videos for training and $48$ for evaluation, and employ the Track AP metric~\cite{russakovsky2015imagenet,yang2019video,dave2020tao} as a proxy measure of the model's ability to capture object permanence.

\vspace{-3mm}
\paragraph{KITTI} is a real-world, multi-object tracking benchmark with city-driving scenarios~\cite{geiger2012we}. The videos are captured at 10 FPS and vary in length. Bounding box annotations are provided for visible parts of the trajectories of people, cars, and a few other categories, but only people and cars are used for evaluation. Following~\cite{tokmakov2021learning}, we split the 21 labeled videos in half to obtain a validation set and use the Track AP metric for evaluation. For completeness, we report test set results with the standard metrics used in this dataset in Appendix~\ref{ap:kitti}.

\subsection{Implementation Details}
Our implementation builds on top of the architecture of~\cite{tokmakov2021learning} and we leave all the components and the hyper-parameters of their model unchanged. Here we only provide the values of the new hyper-parameters. Details of our training and the inference procedures are reported in Appendix~\ref{ap:impl}.
\begin{table*}[bt]
\centering
\small
\addtolength{\tabcolsep}{-5pt}
\caption{Comparison of occluded object localization methods on the test set of LA-CATER using mean IoU. We evaluate on the original version of the benchmark (left) and on the variant with a moving camera (right). Our self-supervised approach outperforms both self- and fully-supervised baselines in almost all scenarios, the gap being especially large on the most challenging Carried task.} \vspace{1mm}
\begin{tabularx}{\linewidth}{l @{\extracolsep{\fill}} C{0.1\linewidth} C{0.1\linewidth} C{0.1\linewidth} C{0.1\linewidth} C{0.1\linewidth} C{0.1\linewidth} C{0.1\linewidth} C{0.1\linewidth}}
\toprule
 &   \multicolumn{4}{c}{LA-CATER Static} & \multicolumn{4}{c}{LA-CATER Moving} \\
\cmidrule(lr){2-5} \cmidrule(lr){6-9}
& Visible   & Occluded         & Contained & Carried           &  Visible   & Occluded         & Contained & Carried      \\
\cmidrule{1-9}
OPNet    & 88.9 &   78.9 &  76.8 & 56.0 & 87.0  & \bf{69.3} & 40.0 & 30.8    \\ \cdashline{1-9}\noalign{\vskip 0.5ex}
OPNet Self.~Sup.  & 89.0 & \bf{81.8}  & 69.0 & 27.5 &	88.0  & 58.0 & 25.8 & 8.5      \\
Heuristic  & 90.1  & 47.0  &  55.4   & 55.9 &	86.7 & 28.6  & 25.4 & 23.3 \\
RAM (Ours) & \bf{91.7}  & 79.3  &  \bf{82.2} &  \bf{63.3} & \bf{90.0} & 62.5  & \bf{55.1} & \bf{51.8}  \\
\bottomrule
\end{tabularx}
\label{tab:la_cater}
\end{table*}

The node embedding head $f_{\mathbf{q}}$ is implemented with a max pooling layer followed by two $1 \times 1$ convolutional layers with with ReLU non-linearities and an L2-normalization layer. We use max pooling layer with kernel $3$ on PD and KITTI and omit pooling on LA-CATER due to low resolution of the frames. The temperature parameter $\tau$ in Equation~\ref{eq:edge} is set to 0.1. We use focal loss~\cite{lin2017focal} when computing the cross entropy in Equation~\ref{eq:ce}. Radius $r$ in Equation~\ref{eq:local} is set to $0.2 \cdot H'$ to balance the representational power of the resulting spatio-temporal graph with computational efficiency. Finally, the individual loss weights $\lambda_{RAM}, \lambda_{over}$ in Equation~\ref{eq:loss} are set to 0.5 and 50 respectively using the validation set of PD.

\subsection{Analysis of Emerging Object Permanence}
In this section, we explore the object permanence hypotheses discovered by our algorithm on LA-CATER. We compare to OPNet - a fully-supervised approach proposed in~\cite{shamsian2020learning}, as well as to their self-supervised variant (referred to as OPNet Self.~Sup.)~and a heuristic-based algorithm from the same paper. The self-supervised baseline is trained to memorize the last visible location of the occluded object, and, the heuristic is designed to predict the target at the center of the closest visible object (presumably, the occluder). Note that, unlike RAM, all these methods operate on pre-computed bounding boxes, and thus detach object permanence reasoning from perception. Experimental results are reported in Table~\ref{tab:la_cater}.

Firstly, we observe that on the original version of the benchmark (left half of Table~\ref{tab:la_cater}) our method achieves top results in almost all scenarios.  In particular, we outperform the self-supervised variant by 35.8 mIoU points in the most challenging Carried category. Their objective assumes that the target remains static once it is not visible. This assumption is very effective for the short-term Occluded scenario, but does not generalize. In contrast, by optimizing for our RAM objective, generic object permanence patterns emerge from the data. For example, our method discovers that once an object becomes a part of another object their locations are tied (see Figure~\ref{fig:lacater}). 
Without explicit supervision it outperforms both the fully-supervised OPNet and the Heuristic which manually encodes this rule by $7$ mIoU points. 
\begin{table*}[bt]
\centering
\small
\addtolength{\tabcolsep}{-5pt}
\caption{Comparison to the state-of-the-art on the validation sets of PD and KITTI using Track AP. All the methods share a detector, tracking algorithm and training data, thus the differences are mainly due to better handling of occlusions. Results for CenterTrack with constant velocity post-processing on KITTI are not reported in prior literature. Our method outperforms both heuristic and learning-based methods in synthetic and real environments despite requiring less supervision.}  \vspace{1mm}
\begin{tabularx}{\linewidth}{l @{\extracolsep{\fill}} C{0.1\linewidth} C{0.1\linewidth} C{0.1\linewidth} C{0.1\linewidth} C{0.1\linewidth} C{0.1\linewidth}}
\toprule
 &   \multicolumn{3}{c}{Parallel Domain} & \multicolumn{3}{c}{KITTI} \\ 
\cmidrule(lr){2-4} \cmidrule(lr){5-7}
& Car AP  & Person AP & mAP & Car AP           &  Person AP   & mAP      \\
\cmidrule{1-7}
CenterTrack    & 66.2 & 54.4 & 60.3 & 77.2 & 51.6  & 64.4   \\
CenterTrack + Const. v.  & 67.6 & 54.9  & 61.2 & - & -  & -     \\
PermaTrack & 71.0  & 63.0  & 67.0   & 84.7 & 56.3 & 70.5   \\
RAM (Ours) & \bf{74.2}  & \bf{69.7}  &  \bf{72.0} &  \bf{87.5} & \bf{64.0} & \bf{75.7}    \\
\bottomrule
\end{tabularx}
\label{tab:track}
\end{table*}

\begin{figure*}
    \centering
    \vspace{-3mm}
    \includegraphics[width=0.95\linewidth]{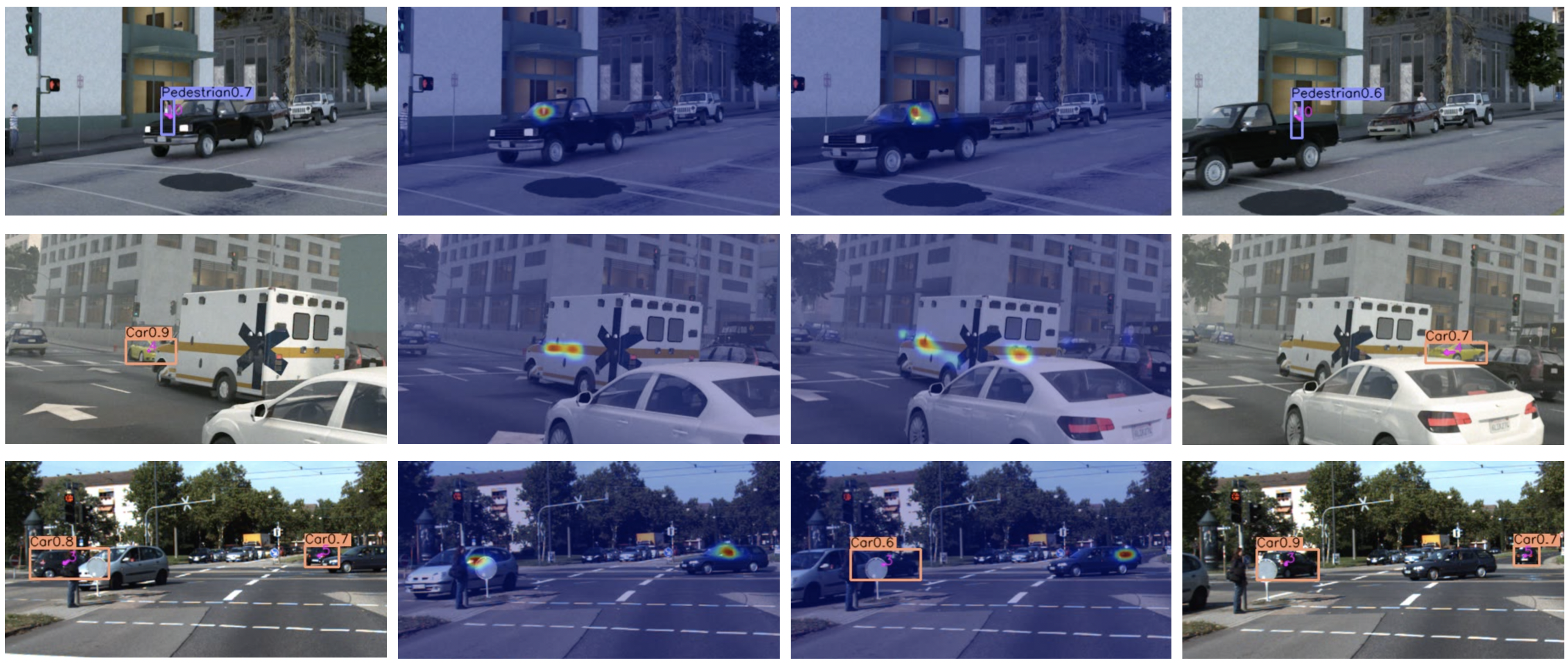}
    \vspace{-4mm}
    \caption{Qualitative results on sequences from the validation sets of PD and KITTI (see \href{https://youtu.be/PGGNyiu4X54}{link} for full results). The model's belief about the location of the invisible objects is visualized with a heatmap overlaid on the frames. Our method forms accurate hypotheses about the locations of occluded objects, resulting in a correct re-identification at the time of disocclusion.}
    \vspace{-4mm}
    \label{fig:real}
\end{figure*}

On the more more challenging version of the dataset with a moving camera (right half of Table~\ref{tab:la_cater}) the performance of all the methods decreases. However, the margins of our approach  increase. The variants that do not use labels for invisible objects (OPNet Self.~Sup.~and Heuristic) encode the fixed camera assumption, which hinders their generalization abilities. In contrast, our method does not make such assumptions and is entirely learned from the data. While the fully-supervised OPNet shows better results, our approach still outperforms it on the most challenging Contained and Carried categories by 15.1 and 21.0 mIoU  respectively.

Finally, we qualitatively analyze object permanence representation learned by our model on the test set of LA-CATER-Moving in Figure~\ref{fig:lacater}. First we show a challenging example of occlusion, where the golden sphere, the red cube and the camera are all in motion. Our model's uncertainty increases, as the scene unfolds. In the end a small part of the object becomes visible, collapsing the walker hypothesis. Next, we show an example of containment, where the target object is consequently covered by brown and blue cones. Despite strong camera motion our model correctly estimates that the sphere remains under the cones. In the last row we can observe another example of double containment followed by carrying. Our approach correctly estimates that as the two cones move the golden sphere moves with them.

\subsection{Capturing Object Permanence in the Real World}
We now demonstrate that our approach is able to discover object permanence patterns in complex, street driving scenes of synthetic PD and real-world KITTI datasets. We compare to CenterTrack~\cite{zhou2020tracking} and PermaTrack~\cite{tokmakov2021learning} which share the same detector architecture, tracking algorithm and training data with our model. Thus, the observed differences reported in Table~\ref{tab:track} are mostly due to the ability of these methods to handle occlusions. In particular, CenterTrack is only trained on visible instances, and has to rely on heuristic post-processing. PermaTrack is a video-based model, which is trained to detect and track invisible objects using explicit supervision in synthetic PD.

We begin our analysis on the PD dataset. CenterTrack is not capable of handling full occlusions and thus serves as a natural baseline. Applying the standard 2D constant velocity heuristic (denoted with 'CenterTrack + Const.~v.' in the table) does result in minor improvements, but overall it is not adequate for driving videos due to strong ego-motion. PermaTrack learns to propagate occluded objects with a constant velocity in the 3D world, compensating for ego-motion changes and thus achieves superior results. Our method outperforms PermaTrack by a significant margin, with improvements being especially noticeable on the more challenging Person category. 
On the real-world KITTI benchmark the observations are similar, confirming that our approach is able to effectively discover object permanence patterns in both synthetic and real environments without explicit supervision or assumptions about object dynamics.

We illustrate the hypotheses about the location of occluded objects discovered by our model on PD and KITTI in Figure~\ref{fig:real}. In the first sequence from the validation set of PD, our model learns to accurately estimate the location of the person occluded by the moving car, resulting in a correct re-identification. In the second sequence, the yellow car starts to turn right before it gets occluded by the ambulance. As the behaviour of the invisible object is unknown to the model, it maintains two distinct hypotheses - one that the car kept moving forward, and another that it continued with the turn. This probabilistic approach allows it to correctly complete the trajectory once the yellow car is revealed. Finally, in the last sequence from the real world KITTI benchmark we illustrate that our model can successfully reason about several occluded objects at a time. 

\subsection{Ablation Analysis}
We conclude by analyzing the importance of various components of our objective on the validation set of PD dataset in Table~\ref{tab:arch_anal}. We use PD for this study due to its balance of scale and realism. Firstly, we observe that the global variant of RAM already achieves comparable results to PermaTrack (row 3 in Table~\ref{tab:track}) without requiring explicit supervision for invisible objects. Using a more computationally efficient form of the objective described in Section~\ref{seq:lrw} (row 2 in the table) does not decrease the performance, confirming our intuition that object motion between consecutive frames is bounded in practice. 
\begin{table}[bt]
\caption{Analysis of the components of our objective using Track AP on the validation set of PD. We ablate graph connectivity, pooling kernel, label smoothing and avoiding center overlaps.}
\vspace{0.15in}
\label{tab:arch_anal}
 \centering
  {
\resizebox{\columnwidth}{!}{
    \begin{tabular}{c@{\hspace{1em}}c@{\hspace{1em}}c@{\hspace{0.5em}}c@{\hspace{1em}}|c@{\hspace{1em}}c@{\hspace{1em}}|c@{\hspace{1em}}}
     Edges & Pool & Smooth & Over. & Car          & Person           & mAP              \\\hline
     Global & $5\times5$ & \xmark & \xmark & 68.9       & 63.5       & 66.2     \\
    
     Local & $5\times5$ & \xmark & \xmark
        & 68.8       & 65.7       & 67.2       \\
     Local & $3\times3$ & \xmark & \xmark
        & 70.1       & 64.8       & 67.9        \\
     Local &$3\times3$  & \checkmark & \xmark
        & 72.7       & 67.8       & 70.2     \\
    Local & $3\times3$  &  \checkmark & \checkmark
        & \textbf{74.2}       & \textbf{69.7}       & \textbf{72.0}    \\
\end{tabular}
}
}
\vspace{-5mm}
\end{table}

Only considering a local node neighborhood during the random walk allows us to decrease the size of the pooling kernel in the node embedding head $f_{\mathbf{q}}$, increasing the feature resolution and thus improving the accuracy of invisible object localization. As we demonstrate in row 3 of Table~\ref{tab:arch_anal}, this translates into higher tracking accuracy due to fewer mistakes during re-identification.

Next, we evaluate the effect of the label smoothing in the RAM objective (see Section~\ref{seq:ws} for details) in row 4 of the table. The relaxed objective is indeed easier to optimize, resulting in a significant performance improvement for both categories. Finally, in the last row of Table~\ref{tab:arch_anal} we demonstrate that overlaps between centers of an occluded object and an occluder are indeed a significant issue and introducing a simple constraint into the objective allows our model to learn to avoid them at inference time, achieving top results. 

For completeness, we have also trained a variant of our model supervising the RAM objective with ground truth locations of invisible instances. It achieves 71.1 mAP points (compared to 72.0 for our best self-supervised variant). Note that the exact location of an invisible object is impossible for a network to predict if the agents' behaviour is non-deterministic. Thus, such labels are noisy from the model's perspective. This result highlights the main advantage of our objective, which implicitly supervises multiple hypothesis about the occluded object location, effectively capturing the non-deterministic nature of the problem.

\section{Conclusion}
Localizing invisible objects is an inherently ambiguous task, making the choice of a learning signal a major challenge. In this work, we proposed a self-supervised objective based on a contrastive random walk along an evolving spatial memory. It is supervised with the ground-truth object centers before and after the occlusion, encouraging the model to store object-centric representations in the memory state, in order to be temporally consistent. We demonstrated that, without additional constraints nor assumptions of dynamics, object permanence patterns emerge in this process. 

\smallsec{Acknowledgements.}
We thank Dian Chen for detailed feedback on the manuscript, and Aviv Shamsian for his help with reproducing OPNet results. AJ was supported by Toyota Research Institute and DARPA MCS.

\bibliography{example_paper}
\bibliographystyle{icml2022}

 \clearpage
\appendix
In this appendix, we provide additional details about our method which were not included into the main paper due to space limitations. We begin by reporting the details of our training and inference procedures in Section~\ref{ap:impl}, further elaborating on the heuristic box refinement strategy used on LA-CATER in Section~\ref{ap:refine}. We then discuss the failure modes of our approach is Section~\ref{ap:fail}. Finally, we conclude by reporting the results of our method on the test set KITTI in Section~\ref{ap:kitti}.

\section{Training and Inference Details}
\label{ap:impl}
Our model is first trained on PD for 28 epochs with with sequences of length 16, exactly following the optimization procedure described in~\cite{tokmakov2021learning}, but only using visible object labels. Note that although Tokmakov et al.~\yrcite{tokmakov2021learning} start from a CenterTrack~\cite{zhou2020tracking} checkpoint pre-trained for 3D object detection on NuScenes~\cite{caesar2019nuscenes}, we have found that starting from an ImageNet or even a self-supervised pre-trained backbone works just as well. We then fine-tune the resulting network on KITTI and LA-CATER. Following~\cite{tokmakov2021learning}, on KITTI we fine-tune the model jointly with PD, but, since both datasets include only visible object labels now, we use KITTI sequences of length 12 during training, in contrast to frame pairs used by~\cite{tokmakov2021learning}. We also find that due to larger resulting batches it is sufficient to fine-tune the model for 3 epochs. 

On LA-CATER we are able to use sequences of length 70 due to the lower resolution of the frames. We only sample sequences that contain occlusion scenarios to speed up convergence. We train the model for 8 epochs with a periodic schedule with step 4, where an epoch is defined as 1000 iterations. Since many occlusion episodes are longer than 70 frames we further fine-tune the model for 2 epochs with a frozen backbone and sequences of length 120. We observe that the objective is harder to optimize on LA-CATER-Moving, thus we double the number of iterations per epoch on that dataset. Finally, we set $r$ to $0.1 * H'$ and $\lambda_{over}$ to 0 on LA-CATER as we notice that this forces the model to more precisely localize the occluded object centers, which is important on this benchmark.

At inference we set the confidence threshold to 0.05 for PD and KITTI and to 0.005 for LA-CATER due to much longer occlusions in that dataset. Following~\cite{zhou2020tracking} we also use a maximum age threshold for an occluded trajectory after which it is terminated. It set to 16 frames for PD and KITTI and to 300 frames for LA-CATER for the same reasons. These values are selected on the validation sets of PD and LA-CATER respectively. On LA-CATER a model is expected to predict precise bounding boxes of invisible objects, however, our approach only estimates an approximate location of the object center. To evaluate on this benchmark we designed a simple, heuristic box refinement algorithm which is described in Appendix~\ref{ap:refine}.

\section{Box Refinement}
\label{ap:refine}
Recall that on LA-CATER the methods are expected to exactly localize invisible objects with a bounding box. However, our algorithm only estimates an approximate location of invisible object centers on a down-sampled feature map $Q \in \mathbb{R}^{D \times H' \times W'}$. A naive approach to converting these centers to boxes is to memorize the size of the object before it was occluded and output a box of the same size around the predicted center. 

In Table~\ref{tab:la_cater_map} we compare our method with this naive box prediction strategy to the baselines from~\cite{shamsian2020learning} on LA-CATER-Moving using a rough mAP@0.1 localization metric. We can observe that without any refinement our approach outperforms all the methods at approximately localizing the invisible objects, with only fully-supervised OPNet being comparable to our approach on the easiest Occluded category. However, our method is not optimized for exactly localizing invisible objects. Firstly, it only provides an approximate location of the object center, as the location of invisible objects is ambiguous beyond the simplest scenarios. Secondly, even this approximate localization is estimated on a down-sampled feature map, which can significantly affect precise localization metrics. To address these issues, we propose a simple, heuristic box refinement strategy.
\begin{table}
\caption{Comparison to the variant of our method without box refinement on the test set of LA-CATER-Moving using mAP@0.1. Our approach outperforms both supervised and unsupervised baselines at approximately localizing invisible objects without post-processing.}
\label{tab:la_cater_map}
  \begin{center}
    \begin{tabular}{l|c@{\hspace{1em}}c@{\hspace{1em}}c@{\hspace{1em}}}
     & Occluded         & Contained & Carried  \\\hline
     OPNet    &    \bf{90.1} &  59.3 & 48.0    \\ \hline
    OPNet Self.~Sup   & 83.8  & 41.7 & 18.7     \\
    Heuristic   & 68.7  &  59.7   & 57.3  \\
    RAM (Ours)  & \bf{90.2}  &  \bf{79.8} &  \bf{76.7}   \\
\end{tabular}
 \end{center}
\end{table}

Our box prediction approach is summarized in Algorithm~\ref{alg:box}. It takes the estimated object track as input, and processes the predictions sequentially. For the frames in which the object is visible, its bounding box is returned without changes. For invisible objects the method uses the center location estimated with our method to infer the box. To this end, we use two flags {\tt was\_moving} and {\tt is\_static}. The former captures whether the object was moving before occlusion and the latter whether it is moving in the current frame. Both are computed based on the changes in the predicted center coordinates between consecutive frames, and do not take camera motion into account. If the object was moving before the occlusion we simply use the naive box estimation strategy discussed above. For the special case of static objects we directly output the last observed bounding box for a more precise localization. Finally, for objects which were static, but are moving in the current frame we refine the bounding box using our proposed heuristic.
\begin{algorithm}
 \caption{\small Bounding box prediction algorithm in Python style. }
 \label{alg:box}
        \definecolor{codeblue}{rgb}{0.25,0.5,0.5}
        \lstset{
          backgroundcolor=\color{white},
          basicstyle=\fontsize{7.2pt}{7.2pt}\ttfamily\selectfont,
          columns=fullflexible,
          breaklines=true,
          captionpos=b,
          commentstyle=\fontsize{7.2pt}{7.2pt}\color{codeblue},
          keywordstyle=\fontsize{7.2pt}{7.2pt},
          escapechar=\&
        }
\begin{lstlisting}[language=python]
for p in track:  # p: model prediction in current frame
  if p["is_visible"]:
    # directly output visible object detections
    out += p["box"]
    last_visible = p["box"]
    continue
  
  # object was moving before occlusion
  if p.was_moving:
    # adjust box to predicted center
    out += get_box(p["center"], last_visible) 
  # object was and remains static
  elif p.is_static:
    # output last visible bounding box
    out += last_visible
  # object was static but is moving now
  else:
    # refine box around predicted center
    out += refine(p["center"], last_visible, others) &\hspace{5mm}& 
\end{lstlisting}
\end{algorithm}

The box refinement heuristic (summarized in Algorithm~\ref{alg:refine}) takes the estimated object center, together with the bounding box of the target object before occlusion and the predicted boxes of the other objects in the current frame (marked with {\tt others}) as input. It makes the assumption that if the center of an invisible object starts moving that must be because it is moving together with the container. To refine the box location it then finds the center of the closest visible box and adjusts the estimated center to be directly below that of the container. Finally, we use the known object size to compute the box around the adjusted center.
\begin{algorithm}
 \caption{\small Bounding box refinement function in Python style. }
 \label{alg:refine}
        \definecolor{codeblue}{rgb}{0.25,0.5,0.5}
        \lstset{
          backgroundcolor=\color{white},
          basicstyle=\fontsize{7.2pt}{7.2pt}\ttfamily\selectfont,
          columns=fullflexible,
          breaklines=true,
          captionpos=b,
          commentstyle=\fontsize{7.2pt}{7.2pt}\color{codeblue},
          keywordstyle=\fontsize{7.2pt}{7.2pt},
          escapechar=\&
        }
\begin{lstlisting}[language=python]
def refine(center, last_visible, others):
  # find the center of the nearest visible object
  closest_center = get_closest(center, others)

  # place the target center under occluder center
  adjusted_center = adjust(center, closest_center)

  # compute the box around adjusted center
  return get_box(adjusted_center, last_visible) &\hspace{5mm}& 
\end{lstlisting}
\end{algorithm}

We now demonstrate that this simple heuristic is sufficient to improve the precise localization results of our method on LA-CATER-Moving in Table~\ref{tab:la_cater_refine}, reporting the strict mean IoU metric. Firstly, we observe that that the proposed algorithm indeed improves our method's localization accuracy on all categories by a significant margin. Secondly, applying the same post-processing step to the fully-supervised OPNet which is trained for precise object localization actually decreases its performance on the Occluded category. On the other two categories OPNet's accuracy improves somewhat, but the margins are smaller and it remains below our self-supervised approach. 
\begin{table}[bt]
\caption{Evaluation of our box refinement strategy on the test set of LA-CATER-Moving using mean IoU. The simple, heuristic algorithm improves the performance of our method, but has mixed results when applied to the fully-supervised OPNet. Overall, the conclusions from the main paper remain unchanged.}
\label{tab:la_cater_refine}
  \begin{center}
    \begin{tabular}{l|c@{\hspace{1em}}c@{\hspace{1em}}c@{\hspace{1em}}}
     & Occluded         & Contained & Carried  \\\hline
    RAM (Ours)  & 55.5  & 40.1 &  34.4   \\
    RAM (Ours) + refine  & 62.5 &  55.1 &  51.8   \\ \hline
    OPNet & 69.3  & 40.0 &  30.8   \\
    OPNet + refine & 64.4  & 46.9 &  41.6   \\
\end{tabular}
 \end{center}
\end{table}

\section{Failure Modes}
\label{ap:fail}
\begin{figure*}
    \centering
    \includegraphics[width=0.9\linewidth]{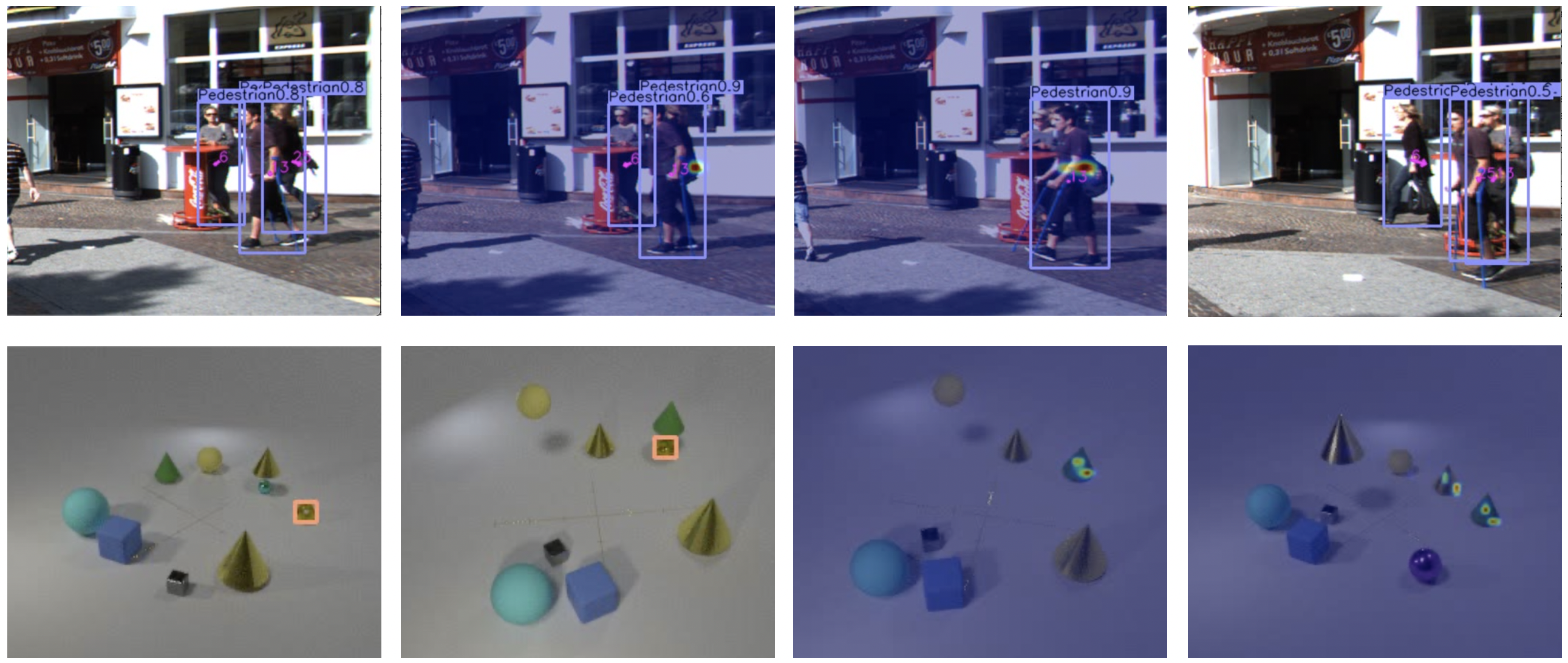}
    \caption{Failure modes of our approach on KITTI and LA-CATER-Moving (see  \href{https://youtu.be/lrrOcS9daMM}{link} for full results). Model’s belief about the location of the invisible objects is visualized with a heatmap overlaid on the frames. Our method suffers from center overlap in the complex, three person occlusion and gets confused when two cones cover two spheres in the direct vicinity from each other.}
    \vspace{-4mm}
    \label{fig:fail}
\end{figure*}

In this section, we discuss the failure modes of our approach, which are shown in Figure~\ref{fig:fail}. In the first example from the validation set of KITTI the centers of the bounding boxes of the three pedestrians overlap in the image plain. As a result, the identities of the three trajectories (shown with numbers in the center of the bounding boxes) switch. Although our overlap avoidance objective improves the method's performance by preventing such mix-ups, it does not always succeed, especially in such challenging cases. 

In the second example from LA-CATER-Moving the green sphere is first covered by the golden cone. Later in the video a different cone covers the golden sphere. At first our method correctly estimates which cone contains the target object, but gets confused over time. 

\section{Evaluation on KITTI Test}
\label{ap:kitti}
Although in this work we did not focus on the problem of multi-object tracking, for completeness we evaluate our approach on the held-out test of the KITTI benchmark, reporting the results in Table~\ref{tab:kitti_test}. We compare to published, vision-based methods using the default metrics for this dataset.  

We observe that on the main HOTA metric~\cite{luiten2020hota} our approach outperforms the state-of-the-art by a significant margin on both categories, despite this metric being less sensitive to accurate object association over occlusions than the Track AP used in the main paper. We note that the MOTA~\cite{bernardin2008evaluating} and MT/PT/ML~\cite{li2009learning} metrics mostly capture object detection accuracy and recall respectively (see~\cite{luiten2020hota} for analysis), and are loosely affected by association accuracy.
\begin{table*}
  \begin{center}
  \resizebox{\linewidth}{!}{
    \begin{tabular}{l|c@{\hspace{1em}}c@{\hspace{1em}}c@{\hspace{1em}}c@{\hspace{1em}}c@{\hspace{1em}}|c@{\hspace{1em}}c@{\hspace{1em}}c@{\hspace{1em}}c@{\hspace{1em}}c@{\hspace{1em}}}
     & \multicolumn{5}{c|}{Car} &
          \multicolumn{5}{c}{Person}\\
    & HOTA $\uparrow$  & MOTA $\uparrow$         & MT $\uparrow$ & PT $\downarrow$  & ML$\downarrow$           & HOTA $\uparrow$ & MOTA $\uparrow$         & MT $\uparrow$  & PT $\downarrow$  & ML $\downarrow$       \\\hline
     SRK~\cite{ODESA2020}   & -  & -  &  - & - & - &  50.9  & 68.0 & 46.4 & 44.7 & \bf{8.9}      \\
    AB3D~\cite{weng2019baseline}  & 69.8  &    83.5  &   67.1   & 21.5 &	11.4 &  35.6  &    38.9 &	17.2 &	41.6 &	41.2 \\
    TuSimple~\cite{choi2015near}   & 71.6  & 86.3  & 71.1 & 22.0 &	6.9 & 45.9  & 57.6 & 30.6 & 44.3 &	25.1   \\
    SMAT~\cite{gonzalez2020smat}   & 71.9  &  83.6  & 62.8 & 31.2 & 6.0 &  - &	- &	-   & - & -  \\
    CenterTrack~\cite{zhou2020tracking}   & 73.0  & 88.8  &   82.2 &  15.4 & 2.5   &  40.4  & 53.8 & 35.4 &  43.3 & 21.3      \\
    DEFT~\cite{Chaabane2021deft_2021_CVPR_Workshops}   & 74.2  & 88.4  &   84.3 &  13.5 & \bf{2.2}   &  -  & - & - &  - & -     \\
    PermaTrack~\cite{tokmakov2021learning} & 78.0  & 91.3  & 85.7 & 11.7 & 2.6 & 48.6  & 66.0 & 48.8 &  35.4 & 15.8 \\ \hline
    RAM (Ours) & \bf{79.5} & \bf{91.6}  &  \bf{86.3} &  \bf{11.2} & 2.5 & \bf{52.7}  & \bf{68.4} & \bf{51.6} &  \bf{34.7} & 13.8  \\

\end{tabular}
}

\caption{Comparison to the vision-based state of the art on the test set of the KITTI benchmark using aggregate metrics. Some methods specialize on a single category. Our generic approach outperforms the state-of-the-art on all metrics except for ML (Mostly Lost) which captures detection recall.}
\label{tab:kitti_test}
 \end{center}
\end{table*}


\end{document}